\documentclass[11pt,a4paper]{article}
\usepackage[hyperref]{naaclhlt2019}
\usepackage{times}
\usepackage{latexsym}
\usepackage{graphicx}  
\usepackage{amsmath} 
\usepackage{booktabs}

\usepackage{url}

\aclfinalcopy 
\def\aclpaperid{1297} 

\title{Improving Human Text Comprehension through Semi-Markov CRF-based Neural Section Title Generation}

\author{Sebastian Gehrmann$^{1,2}$, Steven Layne$^{2,3}$, Franck Dernoncourt$^2$ \\
  $^1$Harvard SEAS, $^2$Adobe Research, $^3$University of Illinois Urbana Champaign \\
  \small{{\tt gehrmann@seas.harvard.edu}}, 
  \small{{\tt solayne2@illinois.edu}},
  \small{{\tt franck.dernoncourt@adobe.com}}}

\date{}

\begin{document}
\maketitle
\begin{abstract}
  Titles of short sections within long documents support readers by guiding their focus towards relevant passages and by providing anchor-points that help to understand the progression of the document. The positive effects of section titles are even more pronounced when measured on readers with less developed reading abilities, for example in communities with limited labeled text resources. 
  We, therefore, aim to develop techniques to generate section titles in low-resource environments. In particular, we present an extractive pipeline for section title generation by first selecting the most salient sentence and then applying deletion-based compression.
  Our compression approach is based on a Semi-Markov Conditional Random Field that leverages unsupervised word-representations such as ELMo or BERT, eliminating the need for a complex encoder-decoder architecture. 
  The results show that this approach leads to competitive performance with sequence-to-sequence models with high resources, while strongly outperforming it with low resources. In a human-subject study across subjects with varying reading abilities, we find that our section titles improve the speed of completing comprehension tasks while retaining similar accuracy. 
\end{abstract}

\section{Introduction}

\noindent Section titles in long documents that explain the content of the section improve the recall of content~\citep{dooling1971effects,smith1992role} while simultaneously increasing the reading speed~\citep{bransford1972contextual}. Additionally, they can provide a context to allow ambiguous words to be understood more easily~\citep{wiley2000effects} and to better understand the overall text~\citep{kintsch1978cognitive}. 
However, most documents do not include titles for short segments or only provide a very abstract description of their topics, e.g.\ ``Geography'' or ``Introduction''. 
This makes them more inaccessible especially to readers with less developed reading skills, who have trouble identifying relevant information in text~\citep{englert2009learning} and therefore more strongly rely on text-markups~\citep{bell2009reading}. 

\begin{figure}[t]
\centering
\fbox{\includegraphics[width=.98\linewidth]{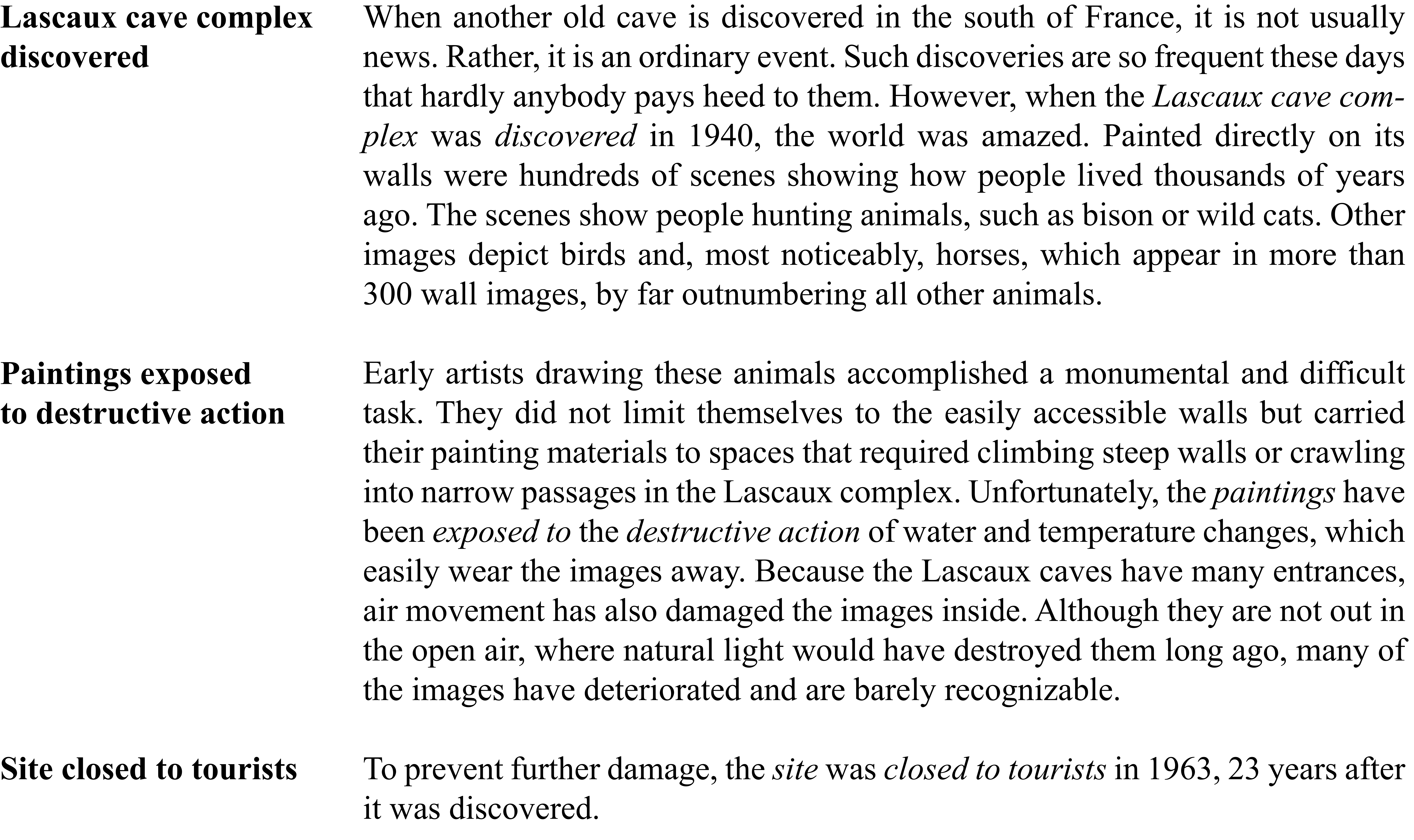}}
\caption{Output example from our model (left) for an out-of-domain text (right).}
\label{fig:teaser}
\end{figure}

This paper introduces an approach to generate section titles by extractively compressing the most salient sentence of each paragraph, as shown in Figure~\ref{fig:teaser}. 
While there has been much recent work on abstractive headline generation from a single sentence~\citep{nallapati2016abstractive}, abstractive models require larger datasets, which are not available in many domains and languages. Moreover, abstractive text-generation models tend to generate incorrect information for complex inputs~\cite{see2017get,wiseman2017challenges}. Misleading headlines can have unintended effects, affecting readers' memory and reasoning skills and even bias them~\citep{ecker2014effects,chesney2017incongruent}. Especially in times of sensationalism and click-baiting, the unguided generation of titles can be considered unethical and we thus focus on the investigation of deletion-only approaches to title-generation. While this restricts this approach to languages that, similar to English, do not lose grammatical soundness when clauses are removed, this approach is highly data-efficient and preserves the original meaning in most cases.  

\begin{figure}[t]
\centering
\includegraphics[width=.98\linewidth]{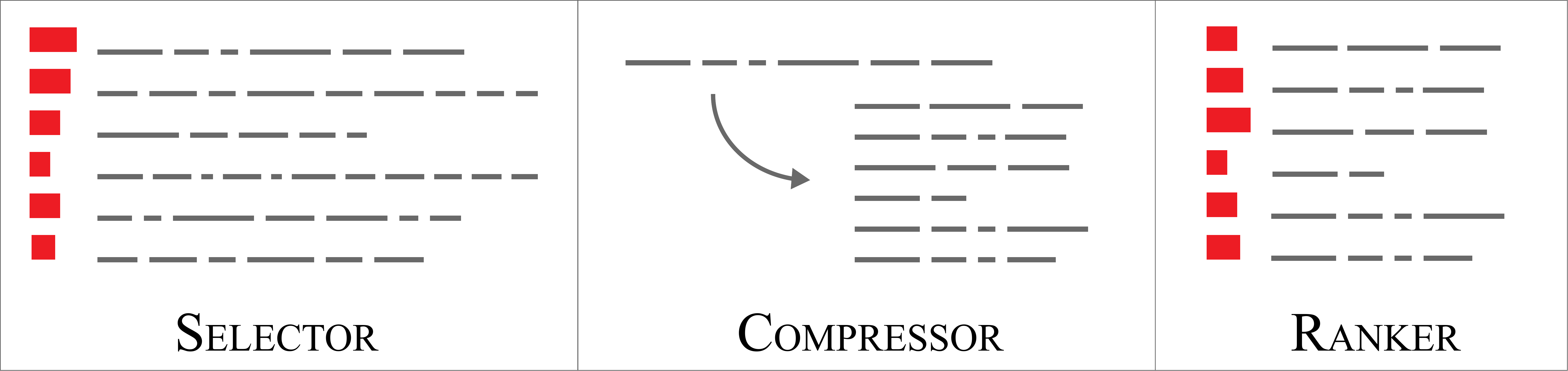}
\caption{Overview of the three steps. The \textsc{Selector} detects the most salient sentence. Then, the \textsc{Compressor} generates compressions and the \textsc{Ranker} scores the them.}
\label{fig:overview}
\end{figure}

We approach the problem with a two-part pipeline where we aim to generate a title for each paragraph of a text, as illustrated in Figure~\ref{fig:overview}. First, a \textsc{Selector} selects the most salient sentence within a paragraph and then the \textsc{Compressor} compresses the sentence. The selector is an extractive summarization algorithm that assigns a score to each sentence corresponding to its likelihood to be used in a summary~\citep{gehrmann2018bottom}. 
Algorithms for word deletion typically rely on linguistic features within a tree-pruning algorithm that identifies which phrases can be excluded~\citep{filippova2013overcoming}. 
Following recent work that shows the efficiency of contextual span-representations~\citep{lee2016learning,peters2018deep}, we develop an alternative approach based on a Semi-Markov Conditional Random Field (SCRF)~\citep{sarawagi2005semi}. The SCRF is further extended by a language model that ranks multiple compression candidates to generate grammatically correct compressions. 

We evaluate this approach by comparing it to strong sequence-to-sequence baselines on an English sentence-compression dataset and show that our approach performs almost as well on large datasets while outperforming the complex models with limited training data. We further show the results of a human study to compare the effects of showing no section titles, human-generated titles, and titles generated with our method. The results corroborate previous findings in that we find a significant decrease in time required to answer questions about a text and an increase in the length of summaries written by test subjects. We also observe that the extractive algorithmic titles have a stronger effect on question answering tasks, whereas abstractive human titles have a stronger effect on the summarization task. This indicates that the inherent differences in how humans and our approach summarize the content of a section play a major role in how reading comprehension is affected. 

\section{Methods}

\subsection{\textsc{Selector}}

To select the most important sentence, we adapt an approach to the problem of content-selection in summarization, which has been shown to be effective and data-efficient~\citep{gehrmann2018bottom}. An advantage of this approach over other extractive summarizers (e.g. \citet{zhou2018neural}) is that those often model dependencies between selected sentences, which is not applicable to this problem since we only aim to extract a single sentence.

Let $x~\textrm{=}~x_1, \ldots, x_n$ denote a sequence of words within a paragraph, and $y~\textrm{=}~y_1, \ldots, y_m$ a multi-sentence summary with $n \gg m$. 
Further, let $t~\textrm{=}~t_1, \ldots, t_n$ be a binary alignment variable where $t_i\textrm{=}1 ~\textrm{iff}~x_i \in y$. Using this alignment, the word-saliency problem is defined as learning a \textsc{Selector} model that maximizes $\log p(t|x) = \sum_{i=1}^{n} \log p(t_i|x)$. Using this model, we calculate the relevance of a sentence $sent := x_{start}, \ldots, x_{end}$, with $1 \leq start < end \leq n$, with a saliency function defined as

\begin{align*}
\textrm{saliency}(sent) &= \frac{1}{|sent|}\sum_{i=1}^{|sent|} p(t_i|sent).
\end{align*}

\noindent The sentence selection problem thus reduces to sentence with the most relevant words within a paragraph $para$,

\begin{align*}
\underset{\text{sent}~\in~\text{para}}{\text{argmax}}\ \textrm{saliency}(sent).
\end{align*}

\noindent We first represent each word using two different embedding channels. The first is a contextual word representation using ELMo~\citep{peters2018deep}, and the second one uses GloVe~\citep{pennington2014glove}. Preliminary experiments corroborated the findings by Peters et al. that the combination of the embeddings help the model converge faster, and perform better with limited training data. 
Both embeddings for a word $x_i$ are concatenated into one vector $e_i$, and used as input to a bidirectional LSTM~\citep{hochreiter1997long}. Finally, the output of the LSTM $h_i$ is used to compute the probability that a word is selected $\sigma(W_s^Th_i+b_s)$ with the trainable parameters $W_s$ and $b_s$.

\subsection{\textsc{Compressor}}

We next define the problem of deletion-only compression of a single sentence. For simplicity of notation, let $x_1, \ldots, x_n$ refer to the words within a single sentence, and $y_1, \ldots, y_n$ be a binary indicator whether a word is kept in the compressed form. The compression $c(x,y)$ becomes 
\[
\textrm{c}(x,y) =
            x_i \forall (x_1,y_1), \ldots, (x_n,y_n)~\text{iff}~y_i = 1.
\]

The challenge here is that choices are not made independently from another. Consider the sentence \emph{The round ball flew into the net last weekend}, which should be compressed to \emph{Ball flew into net}. Here, the choice to include \emph{into} depends on first selecting the corresponding verb \emph{flew}. 
One approach to this problem is to use an autoregressive sequence-to-sequence model, in which choices are conditioned on preceding ones. However, these models typically require too many training examples for many languages or domains. Therefore, we relax the problem by assuming that it obeys the Markov property of order $L$, and train a \textsc{Compressor} model to maximize $p(y|x) = \sum_{i=1}^{n} \log p(y_i|\textbf{x}, y_{i-L:i-1})$. To still retain grammaticality, we define the additional problem of estimating the likelihood of a compression $p(\textbf{y})$ with a \textsc{Ranker} model, described below. 

We compare multiple approaches to deletion-based sentence compression. Throughout, we apply the same embedding as for the \textsc{Selector}. Since a grammatical compression problem relies on the underlying linguistic features~\citep{filippova2013overcoming}, we process the source documents with spaCy~\citep{spacy2} to include the following features in addition to contextualized word embeddings: (1) Part-of-speech tags, (2) Syntactic dependency tags, (3) original word shape, (4) Named entities.
Each of these information is encoded in an additional embedding that is concatenated to the word embeddings. Over the final embedding vector, we use a bidirectional LSTM to compute the hidden representations $h_1, \ldots, h_n$ for this task. 

\paragraph{Naive Tagger} The simplest approach we consider is a naive tagger similar to the \textsc{Selector}. We assume full independence between values in $y$. The probability $p(y_i\text{=}1|x)$ is computed as $\sigma(W_c^T w_1 + b_c)$ with trainable parameters $W_c$ and $b_c$. 

\paragraph{Conditional Random Field} Compressed sentences have to remain grammatical, which implies a dependence between values in $y$. Without any restrictions on the dependence between choices, it is intractable to marginalize over all possible $y$ with a scoring function for a given $(x,y)$ pair,

\begin{align}
    p(y|x) = \frac{\exp \textrm{Score}(x,y)}{\sum_{y'}\exp \textrm{Score}(x,y') }.
    \label{eq:prob}
\end{align}

\noindent Therefore, we assume that only neighboring values in $y$ are dependent, and apply a linear-chain CRF~\citep{lafferty2001conditional} that uses $h$ as its features to this problem. 

We define a scoring function $\textrm{Score}(x, y) = \sum_i \log \phi_i(x,y)$ that computes the (log) potentials at a position $i$ using a function $\phi$. The emission potential for a word $x_i$ is computed as $\phi^{\textsc{e}}_i(x, y) = W_{e2}(\tanh(W_{e1} h_i +b_{e1}))+b_{e2}$, using only the local information. The transition potential $\phi^{\textsc{t}}_i$ depends on the previous and current choice $(y_{i-1}, y_i)$, and can be looked up in a $2 \times 2$ matrix that is learned during training. The complete scoring function can be expressed as

\begin{align*}
    \textrm{Score}(x,y) = \sum_i^{|y|} \left( \phi^{\textsc{e}}_i(x, y) + 
    \phi^{\textsc{t}}_i(y_{i-1}, y_i)\right).
\end{align*}

\noindent During training, we can minimize the negative log-likelihood in which the partition function is computed with the forward-backward algorithm. During inference, this formulation allows for exact inference with the Viterbi algorithm. 

\paragraph{Semi-Markov Conditional Random Field} Although compressed sentences should often include entire phrases, the CRF does not take into account local dependencies beyond neighboring words. Therefore, we relax the Markov assumption and score longer spans of words. This can be achieved with a SCRF~\citep{sarawagi2005semi}. 
Following a similar approach as \citet{Ye-Zhixiu:2018:ACL}, let $s = \{s_1, s_2, \ldots, s_p\}$ denote the segmentation of $x$. Each segment $s_i$ is represented as a tuple $\left<start, end, \tilde{y}_i\right>$, where $start$ and the $end$ denote the indices of the boundaries of the phrase, and $\tilde{y}$ the corresponding label for the entire phrase. To ensure the validity of the representation, we impose the restrictions that $start_1=1$, $end_p=|x|$, $start_{i+1} = end_i+1$, and $start_i \leq end_i$. Additionally, we set a fixed maximum length $L$. 
We extend Eq.~\ref{eq:prob} to account for a segmentation $s$ instead of individual tags such that

\begin{align*}
p(s|x) = \frac{\exp \textrm{Score}(s, x)}{\sum_{s'} \exp \textrm{Score}(s', x)},
\end{align*}

\noindent marginalizing over all possible segmentations. The CRF represents a special case of the SCRF with $L=1$. To account for the segment-level information, we extend the emission potential function to

\begin{align}
\phi^{\textsc{e}}_i(x, \left< \tilde{y}, start, end \right>) = \sum_{i=start}^{end} W_e^T h'_i,
\end{align}

where $h'_i$ is the concatenation of $h_i$, $h_{start} - h_{end}$, and a span length embedding $e_{len}$ to account for both individual words and global segment information. We also extend $\phi^T$ to include transitions to and from longer tagged sequences by representing targets as BIEUO tags~\citep{ratinov2009design}. This formulation allows for similar training by minimizing the negative log-likelihood. 

\subsection{\textsc{Ranker}}

The inference of the CRF and the SCRF require an estimation of the best possible segmentation $s^* =  p(s|x)$, which can be computed using the Viterbi algorithm. However, CRFs and SCRFs are typically employed in sequence-level tagging with no inter-segment dependencies. The sentence compression task differs since a resulting compressed sentence should be grammatical. Therefore, we employ a language model (LM) to rank compression candidates based on the likelihood of the compressed sentence $\textrm{compress}(x,s)$. Namely, we extend the inference target to 

\begin{equation*}
    s^*_{LM} = \underset{s}{\text{argmax}}\  \left( p(s|x) + \lambda p(\textrm{compress}(x,s)) \right),
\end{equation*}

using a weighting parameter $\lambda$. Since exact inference for this target is intractable, we approximate this inference by constraining the re-ranking to the $K$ best segmentations according to a $K$-best Viterbi algorithm.

The \textsc{Ranker} uses the same word embeddings as the \textsc{Compressor}, and a bidirectional LSTM in order to maximize the probability of a sequence $\sum_i p(x_i| x_{1:i-1}) + p(x_i|x_{i+1:n})$. Using the hidden representation $h_i$, we compute a distribution over the vocabulary as $\textrm{softmax}(W_lh_i+b_l)$. We additionally use the same weights for word embeddings and $W_l$, which has been shown to improve language model performance~\citep{inan2016tying}.
We prevent affording an advantage to shorter compressions $c$ during the inference by applying length normalization~\citep{wu2016google} with a length penalty $\alpha$.

\subsection{Sequence-to-Sequence Baselines}
The most common approach to summarization and sentence compression uses sequence-to-sequence (S2S) models that learn an alignment between source and target sequences~\citep{sutskever2014sequence,bahdanau2014neural}. S2S models are autoregressive and generate one word at a time by maximizing the probability $p(y|x) = \sum_i p(y_i|x, y_{1:i-1})$. Since this condition is stronger than that of CRF-based approaches, we hypothesize that S2S models perform better with unlimited training data. However, since S2S models need to jointly learn the alignment and generate words, they typically perform worse with limited data. 

To test this hypothesis, we define two S2S baselines we compare to our models. First, we use a standard S2S model with attention as described by \citet{luong2015effective}. In contrast to the other approaches, this model is abstractive and has the ability to paraphrase and re-order words. We constrain these abilities in a second S2S approach as described by \citet{filippova2015sentence}. This model is a sequential pointer-network~\citep{vinyals2015pointer} and can only generate words from the source sentence. Instead of using an attention mechanism to compute which word to copy, the model enforces a monotonically increasing index of copied words to prevent the re-ordering. We compare both against their reported numbers and our own implementation. 

\section{Data and Experiments}

The \textsc{Selector} is trained on the CNN-DM corpus~\citep{hermann2015teaching,nallapati2016abstractive}, which is the most commonly used corpus for news summarization~\cite{dernoncourt2018summarizationcorpora}. Each summary comprises a number of bullet points for an article, with an average length of 66 tokens and 4.9 bullet points. 
The \textsc{Compressor} is trained on the Google sentence compression dataset~\citep{filippova2013overcoming}, which comprises 200,000 sentence-headline pairs from news articles. The deletion-only version of the headlines was created by pruning the syntactic tree of the sentence and aligning the words with the headline. The largest comparable corpus Gigaword~\citep{rush2015neural} does not include deletion-only headlines.

We limit the vocabulary size to 50,000 words for both corpora. Both \textsc{Selector} and \textsc{Compressor} use a two-layer bidirectional LSTM with 64 hidden dimensions for each direction, and a word-embedding size of 200. Each linguistic feature is embedded into 30-dimensional space.  
During training, the dropout probability is set to 0.5~\citep{srivastava2014dropout}. 
The model is trained for up to 50 epochs or until the validation loss does not decrease for three consecutive epochs. We additionally halve the learning rate every time the validation loss does not decrease for two epochs. We use Adam~\citep{kingma2014adam} with AMSGrad~\citep{reddi2018convergence}, an initial learning rate of 0.003, and a $l2$-penalty weight of 0.001.
The \textsc{Ranker} uses the same LSTM configuration, but we optimize it with SGD with 0.9 momentum, and an initial learning rate of 0.25.

The S2S models have 64 hidden dimensions for each direction of the encoder, and 128 dimensions for the decoder LSTM. They use one layer, and the decoder is initialized with the final state of the encoder. Our optimizer for this task is adagrad with an initial learning rate of 0.15, and an accumulator value of 0.1~\citep{duchi2011adaptive}. 

\subsection{Automated Evaluation}
In the automated evaluation, we focus on the compression models and first conduct experiments with the full dataset to compute an upper bound on the performance of our approach. This experiment functions as a benchmark to investigate how much better the S2S based approaches perform with sufficient data. The next experiment investigates a scenario, in which data availability is limited and ranges from 100 to 1000 training examples. 
We compare results with and without linguistic features to further evaluate whether these features improve the performance or whether contextual embeddings are a sufficient representation. In each experiment, we measure precision, recall, and F1-score of the predictions compared to the human reference, as well as the ROUGE-score. We additionally measure the length of the compressions to investigate whether methods delete a sufficient number of words. 

\begin{figure}[t]
\centering
\includegraphics[width=.95\linewidth]{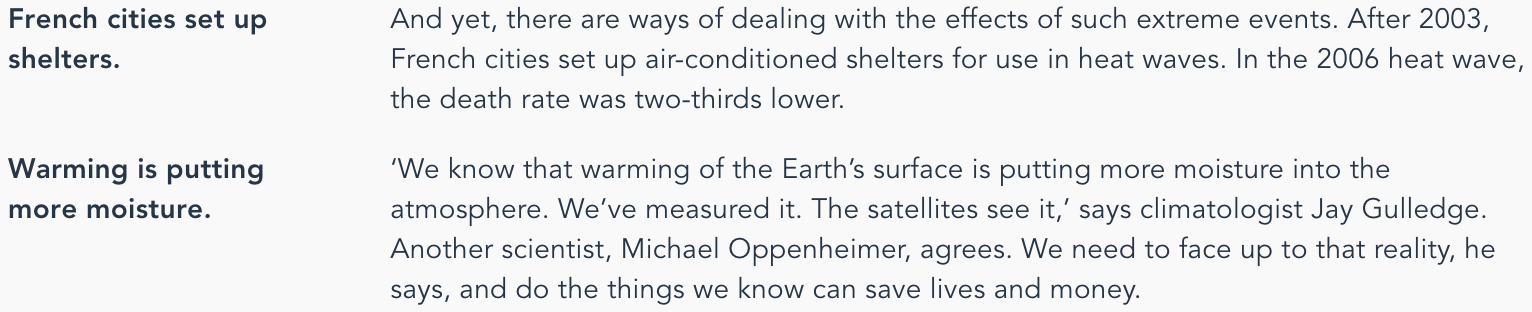}
\caption{Two paragraphs within the interface of our human evaluation with titles in the left-top margin of a paragraph. A side-effect of the data-efficient deletion-only approach, some titles look ungrammatical, as shown in the second example ``\emph{Warming is putting more moisture.}''}
\label{fig:ex}
\end{figure}

\begin{table*}[t]
\centering
\begin{tabular}{@{}llllllllll@{}}
\toprule
Model & Features & \multicolumn{2}{c}{P} & \multicolumn{2}{c}{R} & \multicolumn{2}{c}{F1} & \multicolumn{2}{c}{Length} \\ \midrule

\citet{filippova2015sentence} & Yes &  & &  &  & \textbf{82.0} & &  &  \\ \midrule
S2S w/o copy & No & 83.2 & $\pm$4.5 & 73.0 & $\pm$5.9 & 75.8 & $\pm$4.1 & 9.4 & $\pm$2.9 \\
Sequential Pointer &  & \textbf{87.1} & $\pm$4.1 & \textbf{76.0} & $\pm$5.1 & \textbf{79.1} & $\pm$3.5 & 9.4 & $\pm$2.6 \\
Naive Tagger &  & 81.4 & $\pm$3.7 & 74.6 & $\pm$6.0 & 75.5 & $\pm$3.9 & 9.7 & $\pm$2.7 \\
SCRF &  & 85.2 & $\pm$3.9 & 71.6 & $\pm$8.6 & 73.6 & $\pm$5.5 & 9.0 & $\pm$3.5 \\
SCRF+Ranking &  & 86.1 & $\pm$3.9 & 72.9 & $\pm$8.5 & 74.8 & $\pm$5.5 & 9.1 & $\pm$3.4 \\ \midrule
S2S w/o copy & Yes & 84.6 & $\pm$4.1 & 75.1 & $\pm$5.6 & 77.6 & $\pm$3.8 & 9.5 & $\pm$3.1 \\
Sequential Pointer &  & \textbf{89.6} & $\pm$3.5 & 74.2 & $\pm$5.2 & \textbf{79.8} & $\pm$3.6 & 8.7 & $\pm$2.3 \\
Naive Tagger &  & 84.1 & $\pm$3.5 & \textbf{75.4} & $\pm$5.3 & 79.5 & $\pm$3.7 & 9.5 & $\pm$2.5 \\
SCRF &  & 86.3 & $\pm$3.7 & 73.0 & $\pm$8.4 & 79.1 & $\pm$3.5 & 9.3 & $\pm$2.7 \\
SCRF+Ranking &  & 87.2 & $\pm$3.7 & 73.9 & $\pm$7.8 & 79.6 & $\pm$3.3 & 9.1 & $\pm$2.8 \\ \bottomrule
\end{tabular}
\caption{Results of our models on the large dataset comprising 200,000 compression examples.}
\label{tab:main}
\end{table*}

\subsection{Human Evaluation}

We evaluated the effect of our generated titles in a between-subjects study on Amazon Mechanical Turk. We compared three different conditions: no titles, human-generated titles, and algorithmically generated titles by our SCRF+Ranking model. Every participant kept their randomly assigned condition throughout all tasks. 
We defined the following three tasks to approximately measure the effect of short section titles on (1) retention of text, (2) comprehension of text, and (3) retrieval of information.
\textbf{(Retention)} We first presented a text and then asked participants three questions about facts in the text.
\textbf{(Comprehension)} We showed a text and then asked the participants to generate a three-sentence summary of the text.
\textbf{(Retrieval)} We first presented two questions and then the text, prompting participants to find the answers. 

Previous findings indicate that titles help with retention only when presented towards the beginning of a text~\citep{dooling1973locus}. Thus, we place texts in the left margin at the top of a paragraph as shown in the example in Figure~\ref{fig:ex}. This further avoids interrupting the reading flow of the long text while being integrated into the natural left-to-right reading process. Although reading comprehension is well studied in natural language processing, most datasets focus on machine comprehension~\citep{richardson2013mctest,rajpurkar2016squad}. Therefore, we adapted texts from the interactive reading practice by National Geographic, written by Helen Stephenson\footnote{\url{http://www.ngllife.com/student-zone/interactive-reading-practice}}. The 33 texts are based on articles and comprise three versions for each story; elementary, intermediate, and advanced, from which we selected intermediate and advanced versions. Topics of the texts include Geography, Science, Anthropology, and History; their length ranges from four to seven paragraphs. Each text is accompanied by reading comprehension questions, which we utilized in the retention and retrieval tasks. We first excluded those questions where the answer was part of either human- or algorithmically generated summary. Of the remaining questions, we randomly selected three questions for each of the retention and retrieval tasks. The same questions were shown in either of the conditions.\looseness=-1

Every participant completed six tasks, two for every possible task, one with intermediate and one with advanced difficulty. To account for the different backgrounds of participants, we also asked participants about their perceived difficulty for each task on a 5-point Likert scale. The total time to complete all tasks was limited to 30 minutes, and Turkers were paid \$5. In total, we recruited 144 participants who self-reported that they fluently spoke English, uniformly distributed over the three conditions. They answered on average 68.25\% of questions correctly and took 16.5 minutes to complete all six tasks. This is approximately 30\% faster than the fastest graduate student we recruited for pilot-testing, indicating that Turkers aimed to complete the tasks as fast as possible, possibly by only skimming the text.
We omitted results from participants with an answer accuracy of less than 25\% (n=21), and excluded individual replies given in under 15 seconds (n=10) or over 10 minutes (n=5), leaving a total of 701 completed tasks.
After excluding outliers, the correct answer average was 75.64\%, while the time to completion increased by 15 seconds to 16.75 minutes. 

\section{Results}

\begin{figure}[t]
\centering
\includegraphics[width=.95\linewidth]{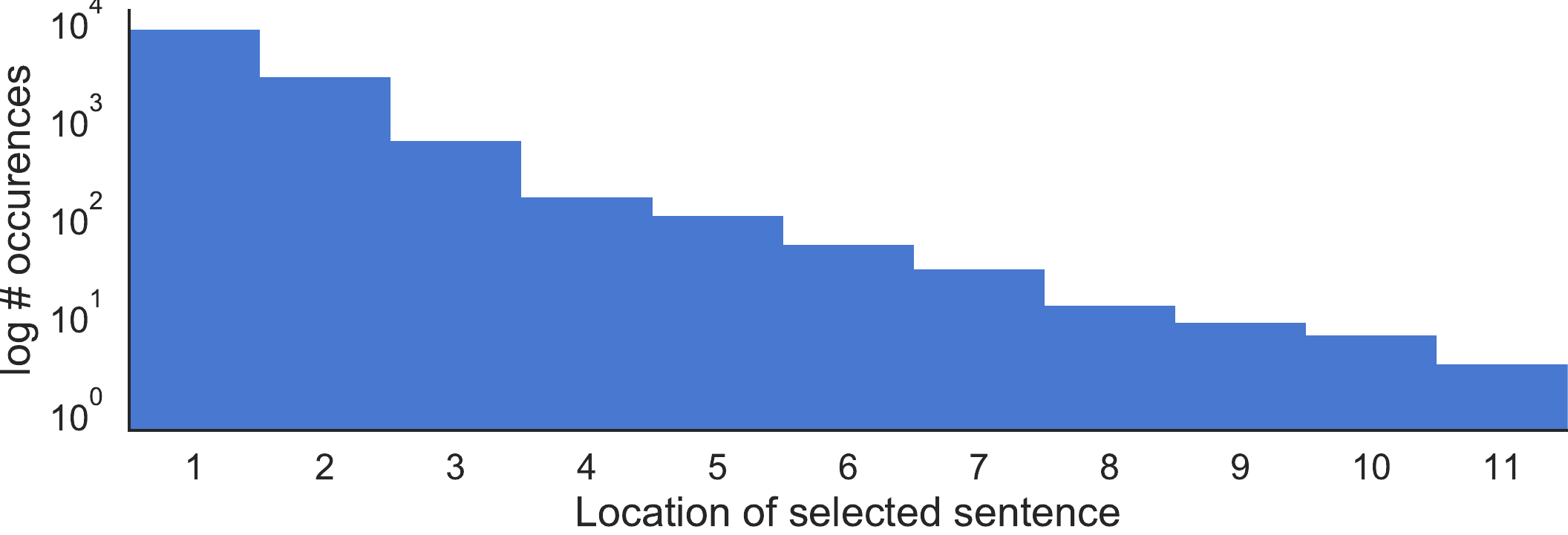}
\caption{Index of extraction within a paragraph.}
\label{fig:loc}
\end{figure}

\paragraph{Selector}
We compare the performance of the selector against the LEAD-1 baseline that naively selects the first sentence of a news article. This provides a strong comparison since a news article typically aims to summarize its content in the first sentence. LEAD-1 achieves ROUGE (1/2/L)-scores of 27.5/9.6/23.7 respectively. In contrast, our selector achieves scores of 30.2/12.2/26.45 which presents an improvement of over 10\% in each category. We illustrate the source of this improvement in Figure~\ref{fig:loc}, which shows the locations of selected sentences and observe a negative correlation between a later location within a text and the probability of being selected. However, in most cases, the first sentence is not the most relevant according to the model. 

\paragraph{Unrestricted Data} 
Table~\ref{tab:main} shows the results of the different approaches on the large dataset. As expected, the copy-model performs best due to its larger modeling potential. It is closely followed by SCRF+Ranking, which comes within 0.2 F1-score when using additional features. This difference is not statistically significant, given the high variance of the results. Compared to the best reported result in the literature by \citet{filippova2015sentence}, our models perform almost as well, despite the fact that their model is trained on 2,000,000 unreleased datapoints compared to our 200,000.
We further observe that all models generate compressed sentences of almost the same length between 8.7 and 9.5 tokens per compression.

The Naive Tagger also achieves comparable performance to the SCRF in F1-Score. To test whether our model leads to a higher fluency compared to it, we additionally measure the \textsc{ROUGE} score. In \textsc{ROUGE-2}, we find that the SCRF+Ranking leads to an increase from 58.1 to 60.1, with an increase in bigram precision by 5 points from 64.5 to 69.3. The Naive Tagger is more efficient at identifying individual words, with a \textsc{ROUGE-1} of 71.3 when the ranking approach only achieves 68.7. While these differences lead to similar \textsc{ROUGE-L} scores between 69.9 and 70.2, the fact that the ranking-based approach matches longer sequences indicates higher fluency. In an analysis of samples from the Naive Tagger, we found that it commonly omits crucial verb-phrases from compressed sentences. 

We show compressions from two paragraphs in the NatGeo data in Figure~\ref{fig:ex}. This example illustrates the robustness of the compression approach to out-of-domain data when including linguistic features. Despite the fact that the example text is not a news article like the training data, it performs well and generates mostly grammatical compressions.

\begin{figure}[t]
\centering
\includegraphics[width=.95\linewidth]{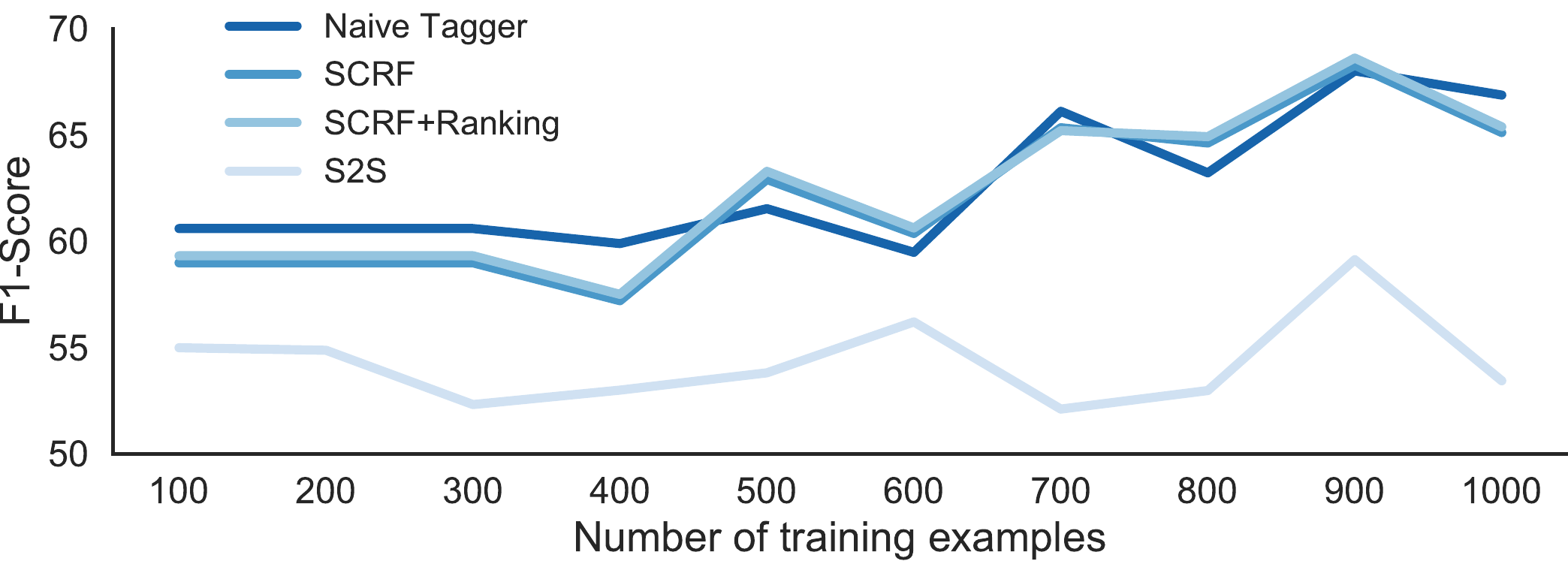}
\caption{F1-scores of the different models with an increasing number of training examples.}
\label{fig:num}
\end{figure}

\paragraph{Limited Data}
We present results on limited data in Figure~\ref{fig:num}. The results show the major advantage of the simpler training objective. All of the tagging-based models outperform the S2S baselines by a large margin due to their data-efficiency. We did not observe a significant difference between the different tagging approaches in the limited data condition. In our experiments, we found that the S2S models start outperforming the simpler models at around 20,000 training examples. Despite its high F1-Score, the Naive Tagger suffers from ungrammatical output in this condition as well, with the readability scoring significantly lower than the SCRF outputs. 
We argue that the SCRF+Ranking approach represents the best trade-off of our presented models since it performs well with limited data while performing almost as well as complex models in unlimited data situations. This makes it most flexible to apply to a wide range of tasks.

\paragraph{Human Evaluation}
In the human study, we notice an immediate effect of the difficulty of texts. Between the intermediate and advanced versions of the texts, the mean time to complete the tasks increases by 8 seconds (from 125 to 133 seconds). Additionally, the mean perceived difficulty increases from 2.40 to 2.55 (in between Easy and Neutral). The largest observed effect of text difficulty is on the accuracy of answers, which decreases from 84.4\% to 67.5\%, indicating that within a similar time frame, the difficult texts were harder to understand. 

\begin{figure}[t]
\centering
\includegraphics[width=.95\linewidth]{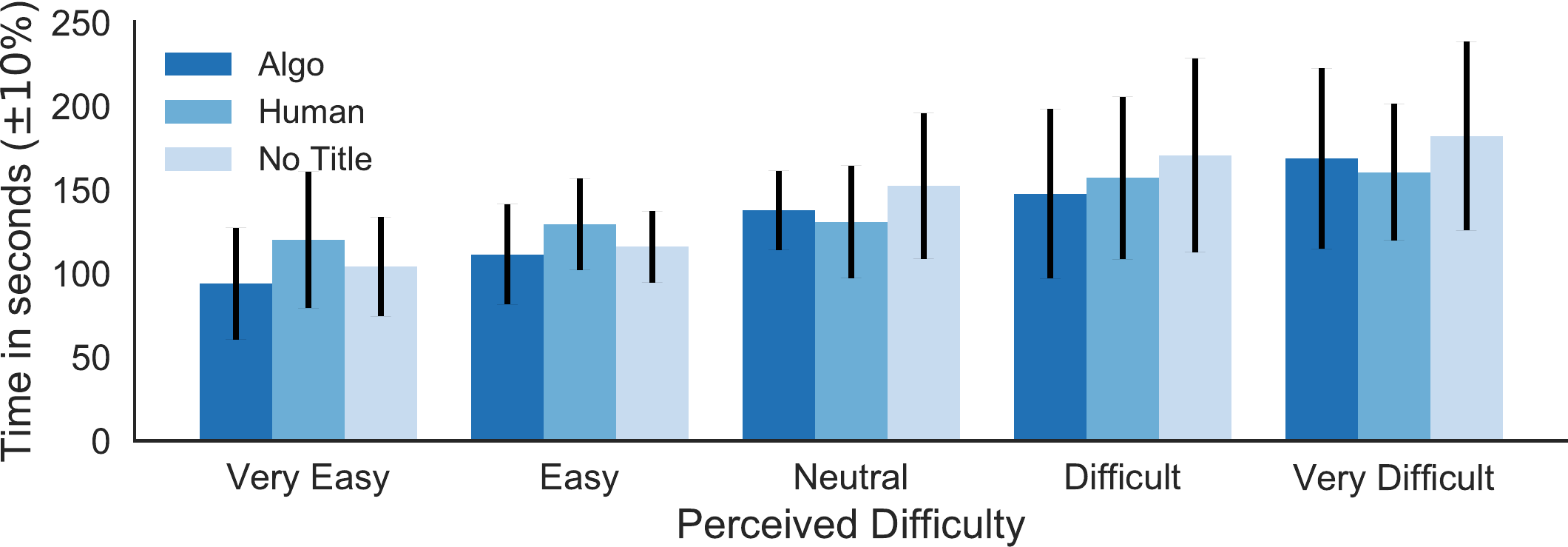}
\caption{Mean and the 90\% confidence interval of the time taken by Turkers to complete the tasks, grouped by the perceived difficulty.}
\label{fig:time}
\end{figure}

\begin{table*}[t]
\centering
\begin{tabular}{@{}lllrr@{}}
\toprule
Task & Measure & Intervention & Effect Size & p-value \\ \midrule
Retention & Time Taken (sec) & Human & -2.2 & 0.63\phantom{*} \\
 &  & Algo & -27.1 & 0.01* \\
 & Accuracy & Human & -0.01 & 0.07\phantom{*} \\
 &  & Algo & -0.01 & 0.13\phantom{*} \\
Retrieval & Time Taken (sec) & Human & -0.9  & 0.87\phantom{*} \\
 &  & Algo & -4.5  & 0.03* \\
 & Accuracy & Human & +0.01 & 0.20\phantom{*} \\
 &  & Algo & -0.01 & 0.15\phantom{*} \\
Comprehension & Time Taken (sec) & Human & -20.9  & 0.03* \\
 &  & Algo & -2.6  & 0.04* \\
 & Summary Length (words) & Human & +8.6  & 0.02* \\
 &  & Algo & +5.3  & 0.03* \\
 & Readability & Human & -0.1 & 0.24\phantom{*} \\
 &  & Algo & -0.1 & 0.50\phantom{*} \\
 & Relevance & Human & -0.02 & 0.92\phantom{*} \\
 &  & Algo & +0.01 & 0.63\phantom{*} \\ \bottomrule
\end{tabular}
\caption{The causal effects of the human and algorithmic section titles on different measures differ across tasks. All the shown effect sizes are measured in comparison to the baseline without any shown titles. Significant p-values at a $0.05$ level are marked with a *.}
\label{tab:human}
\end{table*}

We present a breakdown of time spent on a task by perceived difficulty for each of the conditions in Figure~\ref{fig:time}. There is a positive correlation between perceived difficulty and time spent on a task across all conditions. Interestingly, tasks rated Very Easy and Easy were completed slower in the human condition than in the no-title condition, but faster with the algorithmically generated ones. This effect alleviates in the higher difficulties, in which the no-title condition takes longest. Indeed, a comparison between the algorithmic and no title conditions reveals a decrease in time by 19.8 seconds in the retention task, significant with $p<0.005$ according to a $\chi^2$ test. Interestingly, we can observe opposite effects in the human condition. Here, the comprehension task is completed 15.4 seconds faster ($p<0.005$), but the other tasks only show minor effects. 

To further investigate these effects, we analyze the causal effect of our three conditions by measuring the average treatment effect while controlling for both actual and perceived difficulty of the tasks with an Ordinary Least Squares analysis. Whenever possible, we additionally condition on the total time taken. An overview of our tests is presented in Table~\ref{tab:human}.
In the causal tests, we observe similar effects in the retention task -- the algorithmically generated titles lead to a decrease in time required for the task with an effect size of 27.1 seconds. In contrast, human-generated titles only lead to a non-significant 2.2-second decrease. We observe a non-significant decrease of approximately 4\% in accuracy with added titles. We observed no effect of adding titles on the perceived difficulty of this task. 
In the retrieval task, added titles result in a weaker effect. The algorithmic titles decrease time by only 4.5 seconds, and the human titles by non-significant 0.9 seconds. Similar to the retention task, there is no significant change in accuracy, with all accuracy levels within 1\% from another and we observe no effect on the perceived difficulty.\looseness=-1

\begin{table}[t]
\centering
\begin{tabular}{@{}lrrrr@{}}
\toprule
 & \multicolumn{2}{c}{Readability} & \multicolumn{2}{c}{Relevance} \\ \midrule
Condition & \multicolumn{1}{c}{$\mu$} & \multicolumn{1}{c}{$\sigma$} & \multicolumn{1}{c}{$\mu$} & \multicolumn{1}{c}{$\sigma$} \\
No title & 4.66 & 0.65 & 4.11 & 0.86 \\
Human & 4.55 & 0.76 & 4.09 & 0.95 \\
Algo & 4.52 & 0.72 & 4.12 & 1.02 \\ \bottomrule
\end{tabular}
\caption{Human ratings for human-generated summaries while showing different section titles.}
\label{tab:summ}
\end{table}

In the comprehension task, it is not possible to measure accuracy. Instead, we evaluate readability and relevance as judged by human raters on a five-point Likert scale, two commonly used metrics for abstractive summarization~\citep{paulus2017deep}. We present the average ratings in Table~\ref{tab:summ} and observe that there is almost no difference in relevance and only a minor (not significant) decrease in readability with either condition. 
Curiously, the previous effect on speed reverses in this task -- algorithmic titles only lead to a 2.6-second decrease in time, while human titles lead to a 20.9-second decrease. Both conditions additionally lead to longer summaries; algorithmic titles by 5.3 words and human titles by 8.6 words. 
One potential explanation for this behavior could be that subjects copied the presented section titles into the summary text field. This was not the case, since, on average, only 2.8-3.7\% of the bigrams in the titles were used in the summaries, across both conditions and difficulties (0.6-1.5\% of trigrams, 0.1-0.8\% of 4-grams).  

Given the similar relevance scores, we thus argue that presenting the titles leads to more detailed descriptions of the texts. Similarly to the other tasks, the perceived difficulty does not change significantly. However, we note that some subjects noted that there were some ungrammatical generated titles, which is an artifact of the deletion-only approach. Future work may investigate how abstractive approaches that are not restricted to deletion-based approaches can be applied to the same problem.\looseness=-1

Overall, the results of the human-subject study reveal an effect that is well studied in the literature. Namely, that the type of title influences what is being remembered about a text~\citep{schallert1975improving}, and that different headline styles affect readers in different ways~\citep{lorch2011three}. 
\citet{kozminsky1977altering} found that the immediate free recall of information is biased towards topics emphasized in titles. The better performance in memorization tasks in the algorithmic condition can be explained by the fully extractive approach that immediately shows information judged most relevant by the model. In contrast, human-generated titles show a higher level of abstraction and generalization, which is more helpful for the overall comprehension but does not emphasize any piece of information.

\section{Related Work}

The aim of SCRFs is to learn a segmentation of a sequential input and assigning the same label to an entire segment. While they were originally developed for information extraction~\citep{sarawagi2005semi}, it is most commonly applied to speech recognition within the acoustic model to improve segmentation between different words~\citep{he2015segmental,lu2016segmental,kong2015segmental}. Similar to this work, it has also been shown that coupling an LM with an SCRF can improve segmentation through multi-task training~\citep{lu2017multitask,liu2017empower}. SCRFs have also been applied to sequence tagging tasks, for example, the extraction of phrases that indicate opinions~\citep{yang2012extracting}. 
In this work, we built upon an approach by \citet{Ye-Zhixiu:2018:ACL} who recently introduced a hybrid SCRF that uses both word- and phrase-level information.
Alternative approaches for similar tasks are CRFs that estimate pairwise potentials rather than using a fixed transition matrix~\citep{jagannatha2016structured} or high-order CRFs which outperform SCRFs in some sequence labeling tasks~\citep{cuong2014conditional}.

While this work is the first to apply SCRFs to sentence compression, \citet{grootjen2018automatic} also use extractive summarization techniques to improve reading comprehension by highlighting relevant sentences.
Most similar to our compression approach is Hedge Trimmer~\citep{dorr2003hedge}, which compresses sentences through deletion, but uses an iterative shortening algorithm based on linguistic features. Extending this work, \citet{filippova2013overcoming} apply a similar approach on linguistic features, but learn weights for the shortening algorithm. 
Both approaches also do not consider the selection of the sentence to be compressed, unlike our proposed model. 

\section{Conclusion}

In this work, we have presented a novel approach to section title generation that uses an efficient sentence compression model. We demonstrated that our approach performs almost as well as sequence-to-sequence approaches with unlimited training data while outperforming sequence-to-sequence approaches in low-resource domains. A human evaluation showed that our section titles lead to strong improvements across multiple reading comprehension tasks. Future work might investigate end-to-end approaches, or develop alternative approaches that generate titles more similar to how humans write titles. 

\section*{Acknowledgments}

We are grateful for the helpful feedback from the three anonymous reviewers. We additionally thank Anthony Colas and Sean MacAvaney for the multiple rounds of feedback on the ideas presented in this paper.

\bibliography{naaclhlt2019}
\bibliographystyle{acl_natbib}

\end{document}